\documentclass{article}
\usepackage[ascii]{inputenc}
\usepackage{geometry}
\usepackage{amsmath}
\usepackage{amssymb}
\usepackage{mathtools}
\usepackage[nolist, nohyperlinks]{acronym}
\usepackage{tikz}
\usetikzlibrary{arrows.meta, positioning, calc, fit, backgrounds}
\usepackage{graphicx}
\usepackage{xcolor}
\usepackage{booktabs}
\usepackage{algorithm}
\usepackage[noend]{algpseudocode}
\usepackage{setspace}
\usepackage{authblk}
\usepackage{orcidlink}
\usepackage{hyperref}
\usepackage[
    backend=bibtex,
    bibstyle=numeric-comp,
    sorting=none,
    doi=true]{biblatex}
\newcommand{\ud}{\mathrm{d}}

\graphicspath{{figures/}}
\makeatletter
\def\input@path{{figures/}}
\makeatother

\definecolor{pastel1}{rgb}{0.984313725490196, 0.7058823529411765, 0.6823529411764706}
\definecolor{pastel2}{rgb}{0.7019607843137254, 0.803921568627451, 0.8901960784313725}
\definecolor{pastel3}{rgb}{0.8, 0.9215686274509803, 0.7725490196078432}
\definecolor{pastel4}{rgb}{0.8705882352941177, 0.796078431372549, 0.8941176470588236}
\definecolor{pastel5}{rgb}{0.996078431372549, 0.8509803921568627, 0.6509803921568628}
\definecolor{pastel6}{rgb}{1.0, 1.0, 0.8}
\definecolor{pastel7}{rgb}{0.8980392156862745, 0.8470588235294118, 0.7411764705882353}
\definecolor{pastel8}{rgb}{0.9921568627450981, 0.8549019607843137, 0.9254901960784314}
\definecolor{pastel9}{rgb}{0.9490196078431372, 0.9490196078431372, 0.9490196078431372}

\begin{acronym}
    \acro{MSA}{multiple sequence alignment}
    \acro{ODE}{ordinary differential equation}
    \acro{NeuralODE}[Neural ODE]{Neural Ordinary Differential Equation}
    \acro{MLP}{multilayer perceptron}
    \acro{QKV}{query, key, and value}
    \acro{RK4}{fourth-order Runge-Kutta}
\end{acronym}

\title{Protein Folding with Neural Ordinary Differential Equations}

\author[1,2]{Arielle Sanford\,\orcidlink{0009-0002-5237-8972}\thanks{\href{mailto:arielles@uchicago.edu}{arielles@uchicago.edu}}}
\author[2]{Shuo Sun\,\orcidlink{0009-0006-5775-9730}\thanks{\href{mailto:shuo.sun@tum.de}{shuo.sun@tum.de}}}
\author[2,3]{Christian B.~Mendl\,\orcidlink{0000-0002-6386-0230}\thanks{\href{mailto:christian.mendl@tum.de}{christian.mendl@tum.de}}}
\affil[1]{University of Chicago, Department of Computer Science, 5730 S Ellis Ave, Chicago, IL, 60637 USA}
\affil[2]{Technical University of Munich, School of CIT, Department of Computer Science, Boltzmannstra{\ss}e~3, 85748~Garching, Germany}
\affil[3]{TUM Institute for Advanced Study, Lichtenbergstra{\ss}e~2a, 85748~Garching, Germany}

\date{October 2025}

\begin{document}

\maketitle

\begin{abstract}
Recent advances in protein structure prediction, such as AlphaFold, have demonstrated the power of deep neural architectures like the Evoformer for capturing complex spatial and evolutionary constraints on protein conformation. However, the depth of the Evoformer, comprising 48 stacked blocks, introduces high computational costs and rigid layerwise discretization. Inspired by \acp{NeuralODE}\acused{ODE}, we propose a continuous-depth formulation of the Evoformer, replacing its 48 discrete blocks with a \ac{NeuralODE} parameterization that preserves its core attention-based operations. This continuous-time Evoformer achieves constant memory cost (in depth) via the adjoint method, while allowing a principled trade-off between runtime and accuracy through adaptive \ac{ODE} solvers. Benchmarking on protein structure prediction tasks, we find that the Neural \ac{ODE}-based Evoformer produces structurally plausible predictions and reliably captures certain secondary structure elements, such as $\alpha$-helices, though it does not fully replicate the accuracy of the original architecture. However, our model achieves this performance using dramatically fewer resources, just 17.5 hours of training on a single GPU, highlighting the promise of continuous-depth models as a lightweight and interpretable alternative for biomolecular modeling. This work opens new directions for efficient and adaptive protein structure prediction frameworks.
\end{abstract}

\section{Introduction}
AlphaFold~2 and the improved AlphaFold~3 network have fundamentally transformed protein structure prediction by providing high-confidence atomic-level models for a vast array of proteins, including many without experimentally determined structures~\cite{Alphafold, AlphaFold3, alphafoldimpact}.
This is achieved by combining physical, geometric, and evolutionary information in a deep learning framework that directly predicts 3D structures from amino acid sequences. A central component of AlphaFold~2 is the Evoformer, a deep network trunk composed of 48 non-weight-sharing blocks that progressively refine two representations: the \ac{MSA} and the pairwise residue map. The representations encode domain-specific knowledge, such as geometric reasoning and co-evolutionary coupling, which are essential to AlphaFold's success. However, the model's depth and iterative recycling strategy result in substantial computational cost for both training and inference.

\Acp{NeuralODE} provide an alternative approach for modeling deep transformations~\cite{neuralODE}. Rather than applying a fixed, discrete sequence of layers, a \ac{NeuralODE} parameterizes the derivative of the hidden state using a neural network and integrates these dynamics over a continuous depth variable using a conventional \ac{ODE} solver. This formulation provides several advantages. First, by employing the adjoint sensitivity method for backpropagation, \acp{NeuralODE} achieve constant memory cost with respect to depth, as intermediate activations do not need to be stored during the forward pass. Second, their evaluation strategy can be adaptive. Many \ac{ODE} solvers adjust step sizes dynamically based on the complexity of the trajectory, allowing the computational effort to scale with the difficulty of the input. Third, the trade-off between speed and accuracy becomes tunable, as solver tolerances can be set to balance runtime with numerical precision, making the model adaptable to various deployment scenarios.

Our work is motivated by the observation that the Evoformer's 48 sequential blocks apply small, incremental refinements to the \ac{MSA} and pair representations, which can be viewed collectively as a smooth transformation from their initial to their final state. We propose to replace this discrete stack with a single \ac{NeuralODE} module that models the refinement process as a continuous-time dynamical system. In this formulation, a shared vector field governs the evolution of representations over a continuous depth variable, and an \ac{ODE} solver integrates this field to produce the final state. This substitution reduces the total number of learnable parameters by reusing a single set of weights throughout the integration, and it enables training with constant memory cost via the adjoint sensitivity method. Additionally, the use of \ac{ODE} solvers introduces the potential to control the trade-off between runtime and accuracy through solver tolerances, offering flexibility not available in fixed-depth architectures. For this initial investigation, we omit the recycling steps to isolate the impact of the continuous-depth reformulation. The central assumption of this approach is that a continuous-time model can approximate the cumulative effect of the 48 Evoformer blocks, which is reasonable given that each block contributes a small, progressive update to the overall representation.

In this work, we use OpenFold~\cite{Openfold} to generate reference data, and treat its Evoformer transformation as ground-truth. OpenFold is an open-source implementation of AlphaFold~2 that reproduces the original architecture and performance while offering greater flexibility for model customization. OpenFold provides access to the full model architecture, eliminating dependence on proprietary code or Google infrastructure. All training data used in this study, namely, the \ac{MSA} and pairwise residue representations in various Evoformer blocks, were derived from OpenFold inference runs on monomer inputs, using a single pass without recycling. These representations are treated as ground truth trajectories that define the transformation our \ac{ODE} must learn to emulate. 

Our implementation is available online at \cite{openFoldODE}.

\section{Background}

\subsection{AlphaFold~2 Architecture}

The Evoformer, the key building block of AlphaFold~2, is a deep network trunk composed of 48 non-weight-sharing blocks that progressively refine the \ac{MSA} and the pairwise residue map representations. Each block applies Transformer-style updates, rooted in the attention mechanisms introduced by Vaswani et al.~\cite{attention}, including gated axial attention over the \ac{MSA} and geometry-aware updates to the pairwise representation via triangle-based modules that capture interactions among residue triplets.

\subsection{\Acp{NeuralODE}}

A widely used ingredient of artificial neural networks is a \emph{residual block} transformation
\begin{equation}
\label{eq:residual_block}
s_{t+1} = s_{t} + f(s_t, \theta_t),
\end{equation}
with $s_t \in \mathbb{R}^D$ the (hidden) state at layer (or time step) $t$, and $f$ a trainable function parametrized by $\theta_t$. \Acp{NeuralODE} \cite{neuralODE} are motivated by the insight that Eq.~\eqref{eq:residual_block} can be interpreted as an Euler step discretization of an \ac{ODE}. Taking the limit of infinitely small time steps, one arrives at the continuous-time analogue
\begin{equation}
\frac{\ud s(t)}{\ud t} = f(s(t), t, \theta).
\end{equation}
This formulation opens the door to a plethora of well-studied numerical \ac{ODE} solvers and time step adaptation. Backpropagation can be implemented memory-efficiently via the ``adjoint sensitivity method'', i.e., by solving an auxiliary \ac{ODE} that runs backward in time.

\section{Evoformer \Ac{ODE} Architecture}

To create a continuous-depth analog of AlphaFold's Evoformer, we define an \ac{ODE} function that approximates the cumulative transformation of the \ac{MSA} and pair representations over depth. This function is inspired by the core operations of a single Evoformer block and retains the bidirectional exchange of information between the \ac{MSA} and pairwise residue representations

The function accepts the current \ac{MSA} and pair states \((m, z)\) and returns their time derivatives \((\frac{dm}{dt}, \frac{dz}{dt})\). Internally, it computes one full pass of Evoformer-style updates to obtain updated states \((m', z')\), then compares these to the inputs to estimate a continuous vector field. The difference between the updated and original states is scaled by time-dependent gating functions \(\sigma_m(t)\) and \(\sigma_z(t)\), learned via a shallow \ac{MLP} over the scalar depth variable \(t\), resulting in a smooth evolution of the representations along the depth trajectory:
\begin{align}
f(m, z) &= \big(\sigma_m(t) \cdot (m' - m), \sigma_z(t) \cdot (z' - z)\big),\\
\frac{\ud (m, z)}{\ud t} &= f(m, z)
\end{align}
where \(\sigma_m(t)\) and \(\sigma_z(t)\) act as mixing coefficients that modulate the magnitude of updates throughout integration.

\begin{figure}[ht]
  \centering
  \begin{tikzpicture}[
    box/.style={font=\scriptsize, draw, rounded corners, minimum height=30, minimum width=37, align=center, inner sep=2},
    legend/.style={font=\scriptsize, draw, align=center, inner sep=4},
    arrow/.style={-stealth, thick},
    timearrow/.style={-stealth, thick, dashed},
    odesolverarrow/.style={-stealth, thick, draw=pastel8!20!purple}
]

\node[box, fill=pastel1] (msa_in) at (-6, 1) {MSA\\(s, r, c)};
\node[box, fill=pastel2] (pair_in) at (-6, -1) {Pair\\(r, r, c)};


\node[box, fill=pastel5] (row_attn) at (-4, 1) {Row-wise\\self-attn.};
\node[box, fill=pastel5] (col_attn) at (-2.25, 1) {Col-wise\\attn.};
\node[box, fill=pastel3] (msa_trans) at (-0.5, 1) {MSA\\transition};

\node[box, fill=pastel4] (opm) at (-0.5, -1) {Outer\\product\\mean};

\node[box, fill=pastel6] (tri_up) at ( 1.25, -1) {Triangle\\update};
\node[box, fill=pastel3] (pair_trans) at ( 3, -1) {Pair\\transition};

\draw[arrow] (row_attn) -- (col_attn);
\draw[arrow] (col_attn) -- (msa_trans);
\draw[arrow] (msa_trans) -- (opm);
\draw[arrow] (opm) -- (tri_up);
\draw[arrow] (tri_up) -- (pair_trans);

\begin{pgfonlayer}{background}
\draw[
  fill=gray!30,      
  draw=none,         
  rounded corners
] (-5,-1.85) rectangle (5, 1.85);
\node at (0, 2.1) {\scriptsize $f(\mathrm{MSA}, \mathrm{Pair}, t, \theta)$};
\end{pgfonlayer}


\draw[arrow] (msa_in) -- (row_attn);
\draw[arrow] (pair_in) -- (opm);
\draw[arrow] (-4, -1) -- (row_attn);

\node[box, fill=pastel8] (dmsa) at (6, 1) {$\dfrac{\ud}{\ud t}\mathrm{MSA}$};
\node[box, fill=pastel8] (dpair) at (6,-1) {$\dfrac{\ud}{\ud t}\mathrm{Pair}$};

\draw[timearrow] (msa_trans) -- (dmsa) node[midway, above]{$\sigma_m(t)$};
\draw[timearrow] (pair_trans) -- (dpair) node[midway, above]{$\sigma_z(t)$};

\draw[odesolverarrow]
  (dmsa.north) .. controls ++(0, 2) and ++(0, 2) .. (msa_in.north);

\draw[odesolverarrow]
  (dpair.south) .. controls ++(0,-2) and ++(0,-2) .. (pair_in.south);

\node[legend, anchor=south] at (0,3.5) (legend) {
  \raisebox{0.5mm}{\tikz{\draw[arrow] (0,0) -- (0.6,0);}} state flow \quad
  \raisebox{0.5mm}{\tikz{\draw[timearrow] (0,0) -- (0.6,0);}} time scaling \quad
  \raisebox{0.5mm}{\tikz{\draw[odesolverarrow] (0,0) -- (0.6,0);}} ODE solver
};
\end{tikzpicture}
  \caption{\Ac{ODE} Evoformer block. Arrows show the information flow. The shape of each array is shown in parentheses, where $s$ is the number of \ac{MSA} sequences, $r$ is the number of residues, and $c$ is the channel dimension.}
  \label{fig:ode_evoformer}
\end{figure}
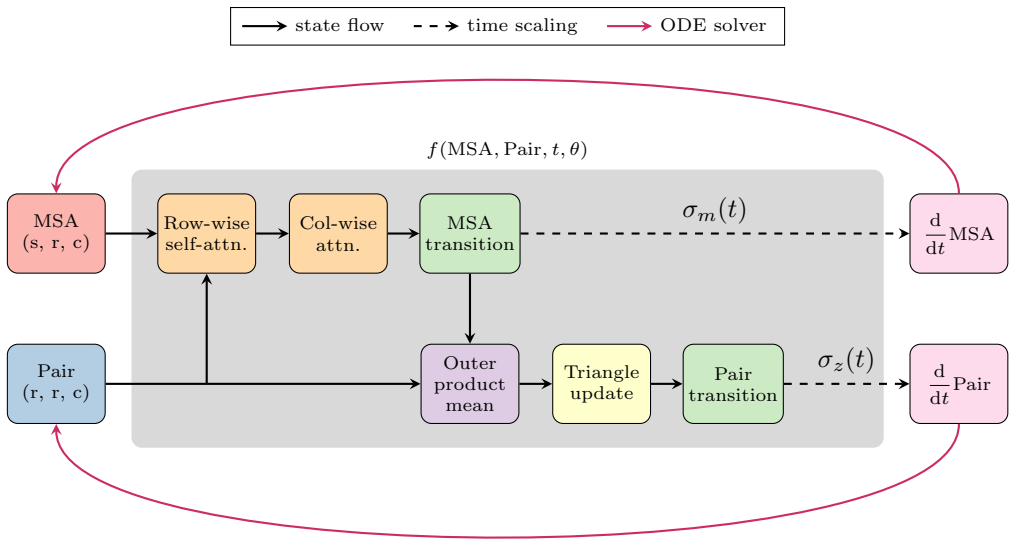

The architecture is visualized in Fig.~\ref{fig:ode_evoformer}. In detail, the model applies the following sequence of transformations to the input representations:
\begin{itemize}
    \item \textbf{\ac{MSA} row-wise attention with pair bias:} Applies multi-head self-attention over residues, incorporating pairwise residue information as an additive bias. This models long-range interactions across residue positions.
    \item \textbf{\ac{MSA} column-wise attention:} Performs attention across sequences for each residue position, refining the evolutionary context of the alignment.
    \item \textbf{\ac{MSA} transition:} A feedforward network applied pointwise to the \ac{MSA}, modeling intra-residue nonlinearity.
    \item \textbf{Outer product mean:} Projects \ac{MSA} embeddings to lower-rank vectors and computes an outer product averaged across sequences, updating the pairwise representation.
    \item \textbf{Triangle update (multiplicative):} Captures higher-order residue interactions by updating each pair $(i, j)$ through multiplicative paths involving a third residue $k$, modeling triangular relationships between residues in the structure.
    \item \textbf{Pair transition:} A residual \ac{MLP} applied to the pair representation for local refinement.
\end{itemize}

The \ac{ODE} function reimplements core Evoformer operations in a simplified form to reduce memory usage and enable integration into a continuous-depth architecture, while preserving the domain-specific assumptions that make the Evoformer effective for protein structure modeling. To reduce overhead, we omit both triangle attention modules (incoming and outgoing), which, in AlphaFold, use self-attention to propagate pairwise features along triangular paths and are among the most memory-intensive components. Our \ac{MSA} attention layers remove chunked softmax, fused kernels, and advanced bias handling, instead using straightforward \ac{QKV} projection and gating. The outer product mean computes a reduced-rank elementwise interaction rather than AlphaFold's full outer product. The triangle multiplication update, which AlphaFold splits into separate incoming and outgoing branches, is replaced with a symmetric approximation. The pair transition module is implemented as a basic two-layer \ac{MLP} with ReLU and normalization, omitting the chunked execution used in AlphaFold's inference stack. These design choices significantly reduce the computational footprint, enabling the entire Evoformer-like transformation to be modeled as a single continuous vector field. Rather than stacking 48 discrete blocks, we model representation refinement as a continuous process over depth, allowing the internal state to evolve smoothly under numerical integration. Appendix~\ref{sec:evoformer_ode_details} lists and describes the trainable parameters and dimensions of the architecture in detail.

\section{Training}

We trained the Evoformer \ac{ODE} model in two distinct phases using a curated dataset of protein monomers. The dataset comprises 500 proteins for training (reduced to 370 after applying a sequence length filter of 350 residues), along with 50 validation proteins and 50 test proteins. The proteins were selected to span a diverse range of sizes, functions, and organisms, ensuring robustness and generalization. For each protein, we retrieved the primary sequence and its \ac{MSA} from the OpenProteinSet~\cite{openproteinset_paper, openproteinset}.
Structural templates from the PDB70 database~\cite{mmseqs} were incorporated during OpenFold inference to provide coarse-grained structural priors based on sequence similarity to known proteins. All three inputs, the primary sequence, the \ac{MSA} and the structural templates, are required to run full OpenFold inference, which we used to generate the training data. To train the continuous-depth Evoformer, we saved the \ac{MSA} and pair representations at both the entry and exit of the 48-block Evoformer, as we ran OpenFold inference on a multitude of proteins, capturing the transformation that the OpenFold Evoformer performs on these features. For a subset of proteins, we additionally saved intermediate representations at each block, creating a dense trajectory of the Evoformer's internal evolution. These data provide direct supervision for learning the continuous vector field that governs the transformation of the representations over depth.

The training process consists of a preliminary phase and a main phase, each with distinct goals. The preliminary phase is designed to expose the model to the natural dynamics of AlphaFold's Evoformer. It was conducted on a subset of 80 proteins and supervised using intermediate representations sampled from within the 48-block Evoformer trunk at regular strides (blocks 0, 8, 16, 24, 32, 40, 48). By learning from partial refinement trajectories rather than only final states, this phase encourages the model to capture the incremental evolution of representations through the folding process.

The main training phase focuses on learning the complete transformation from the initial to the final state. Using the full set of 500 training proteins, the model is trained to predict the output of the 48th Evoformer block given only the input to the first block. This supervision forces the \ac{NeuralODE} to learn a vector field that integrates over the entire folding trajectory. In this phase, intermediate blocks are not used, and only the endpoints of the transformation are involved in the loss computation.

We adopted several training configurations to improve computational efficiency at the cost of some predictive accuracy. The best-performing model was trained for 17.5 hours, at which point it reached the tolerance threshold for validation loss improvement. The internal hidden dimensionality used across all attention and feedforward submodules was set to 64, applied uniformly to both the \ac{MSA} and pair representations. In contrast, AlphaFold utilizes separate hidden dimensions for each stream: 384 for the \ac{MSA} representation ($c_m$) and 128 for the pair representation ($c_z$), thereby increasing its capacity to model complex interactions at the expense of higher memory and computational requirements. Further, we reduced the number of \ac{MSA} sequences passed to the model to a clustered subset of at most 64, limiting the effective \ac{MSA} depth during attention-based updates. This is taken off the top half of the 128 clustered sequences in Openfold's \ac{MSA} representation, which are selected via sequence clustering algorithms to remove redundancy while preserving evolutionary signal. We used the classical \ac{RK4} method as the \ac{ODE} solver, as opposed to more accurate but computationally expensive solvers, such as the Dormand-Prince method (DOPRI5). The maximum allowed sequence length was capped at 350 residues, and in the preliminary training phase, intermediate representations were sampled every fourth block. Preliminary training was limited to a maximum of 20 epochs, while main training was bounded by the tolerance.

We ran the whole training process on a single workstation computer with the following hardware configuration:  NVIDIA Quadro P4000 GPU (8~GB memory), Intel Core i7-7700 CPU at 3.60~GHz, 4 cores (8 HW threads), and 32~GiB DDR4 main memory.

\section{Results}

\begin{figure}[h!]
    \centering
    \includegraphics[width=0.8\textwidth]{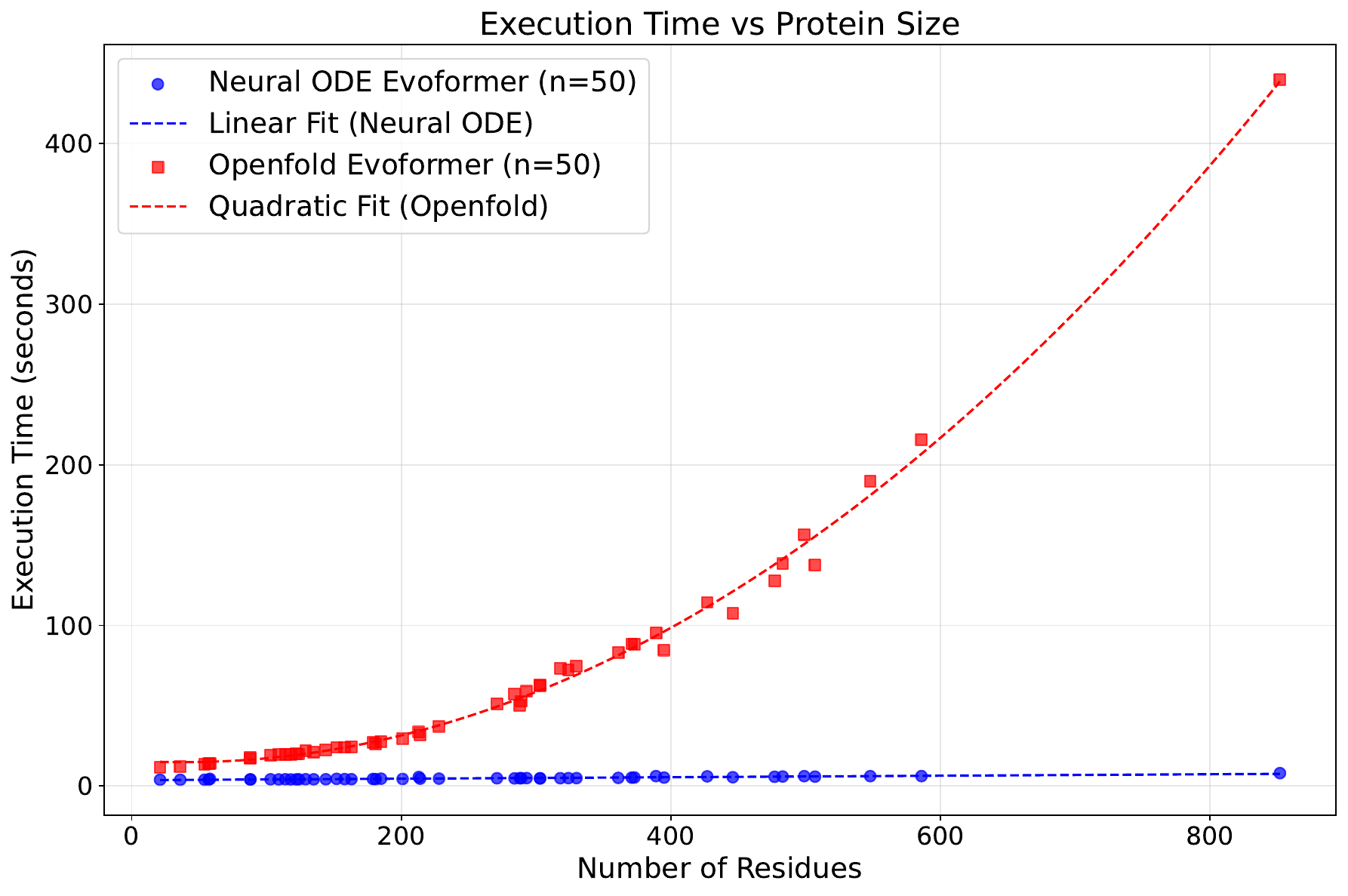}
    \caption{Inference runtime as a function of protein length. Linear and quadratic fits are shown for the \ac{NeuralODE} and OpenFold Evoformer, respectively.}
    \label{fig:runtime_plot}
\end{figure}

We evaluated the inference speed of our \ac{NeuralODE}-based Evoformer implementation on 50 test proteins and compared it against Openfold's standard Evoformer stack. Both models were tested on the same set of sequences, ranging from 21 to 852 residues. A benchmark plot illustrating the relationship between sequence length and total runtime is shown in Figure~\ref{fig:runtime_plot}.

Quantitatively, the \ac{NeuralODE} Evoformer achieved an average runtime of 4.85 seconds per protein, corresponding to approximately 0.0300 seconds per residue. In contrast, OpenFold's Evoformer required an average of 65.06 seconds per protein, or about 0.2230 seconds per residue --- over 7 times slower per residue on average. The fitted runtime trends show a linear scaling for our model:
\[
\text{\ac{NeuralODE} fit: } \text{time} = 0.004625 \times \text{residues} + 3.64
\]
versus a quadratic dependence for OpenFold:
\[
\text{OpenFold fit: } \text{time} = 0.000641 \times \text{residues}^2 - 0.0497 \times \text{residues} + 15.83
\]
These results demonstrate that replacing the 48-block Evoformer with a continuous-depth \ac{NeuralODE} significantly reduces computational cost, especially for long proteins, while preserving critical structure prediction capabilities.

\begin{figure}[h!]
    \centering
    \includegraphics[width=0.95\textwidth]{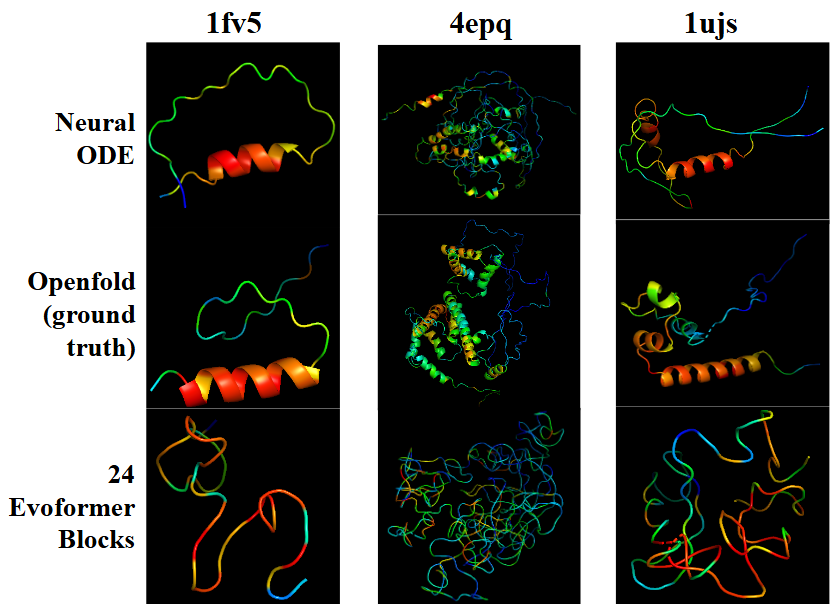}
    \caption{Predicted protein structures visualized using PyMOL~\cite{pymol}. Columns correspond to different proteins (\texttt{1fv5}, \texttt{4epq}, \texttt{1ujs}). Rows show predictions from (top) the \ac{NeuralODE} Evoformer, (middle) OpenFold ground truth, and (bottom) a model using only 24 Evoformer blocks. Color denotes per-residue confidence (pLDDT; red = high, blue = low).}
    \label{fig:structure_pymol}
\end{figure}

To assess the structural quality of predictions generated by the \ac{NeuralODE} Evoformer, we visualized predicted protein structures for three examples: \texttt{1fv5}, \texttt{4epq}, and \texttt{1ujs}. Figure~\ref{fig:structure_pymol} compares the predicted structures from three configurations: (1) our \ac{NeuralODE} Evoformer, (2) the full OpenFold model (used as ground truth reference), and (3) a truncated OpenFold model using only 24 Evoformer blocks.

The \ac{NeuralODE} Evoformer captures several key features of the full Evoformer's output. In particular, $\alpha$-helices are generally identified and positioned correctly, and the overall folding behavior often reflects the major global structure. However, there are clear deviations from the ground truth in many fine-grained details --- especially in loop regions and global packing. The differences are more pronounced for large or complex folds such as \texttt{4epq}. Compared to the 24-block truncated Evoformer, however, the \ac{ODE}-based model produces substantially more organized structures with higher-confidence predictions.

Overall, these examples suggest that the \ac{NeuralODE} Evoformer, although not replicating the full Evoformer output, can capture broad structural features, such as secondary structure and global topology. This provides a promising proof of concept for reducing memory and runtime costs while maintaining biologically relevant predictions.

\section{Conclusions and Next Steps}

This work demonstrates that a continuous-depth reformulation of AlphaFold~2's Evoformer is both feasible and efficient. By replacing the original 48 discrete Evoformer blocks with a single \ac{NeuralODE} module, we achieve substantial runtime improvements while maintaining reasonable agreement with the structural patterns produced by the original model. The fact that this result was achieved with only 17.5 hours of training on a single GPU underscores the effectiveness and potential of the approach.

In stark contrast, AlphaFold's original model was trained over 11 days using hundreds of TPU v3 cores, with a compute cluster size equivalent to 512 GPUs and access to vast curated datasets. Given these constraints, our results demonstrate that continuous-depth approximations can effectively represent the behavior of large-scale models with significantly fewer resources.

We deem the following points natural extensions and promising ideas for future work:
\begin{itemize}
\item \emph{Larger \ac{MSA} cluster size:} Increase the number of clustered \ac{MSA} sequences used per input from 64 to 128 or more to match AlphaFold's diversity-preserving input.
\item \emph{Increased hidden dimensions:} Use distinct and larger hidden sizes for each representation, such as $c_m = 384$ and $c_z = 128$, as in AlphaFold.
\item \emph{Improved \ac{ODE} solver:} Replace the fixed-step \ac{RK4} solver with an adaptive solver such as Dormand-Prince for better numerical stability and precision.
\item \emph{Uncapped sequence length:} Remove the 350-residue sequence length cap to allow full-length protein modeling.
\item \emph{Denser supervision in pretraining:} Replace stride-4 block sampling with full per-block supervision during the preliminary training phase.
\item \emph{Flexible epoch schedule:} Remove the fixed 20-epoch limit in preliminary training and instead rely on validation loss dynamics.
\item \emph{Larger and more diverse training set:} Improve generalization in both the preliminary and main training stages.
\end{itemize}

In conclusion, our research provides an encouraging proof of concept: that continuous-time modeling can serve as a lightweight and scalable alternative to deep stacked architectures in protein structure prediction. With access to greater resources, the \ac{NeuralODE} Evoformer has the potential to match the performance of traditional block-based models while offering unique advantages in flexibility, memory usage, and theoretical interpretability.

\section*{Acknowledgements}

A.~Sanford gratefully acknowledges support from the Rice University Wagoner Foreign Study Scholarship, which enabled the international research stay during which this project was conducted. S.~Sun thankfully acknowledges funding support from the BMW Group.

\printbibliography

\newpage

\appendix

\section{Details of the Evoformer ODE Architecture}
\label{sec:evoformer_ode_details}

We provide more technical details of the Evoformer ODE architecture in the following sections, including the dimensions of the trainable tensors and the algorithms employed. The source code of our implementation is available at \cite{openFoldODE}.

\subsection{Notation and Shapes}

Let $m \in \mathbb{R}^{S \times R \times c_m}$ denote the multiple sequence alignment (MSA) representation, and let
$z \in \mathbb{R}^{R \times R \times c_z}$ denote the pairwise residue representation. The scalar depth variable is $t \in [0,1]$. The integer variables are defined below.

\begin{center}
\begin{tabular}{@{}ll@{}}
\toprule
Symbol & Meaning \\
\midrule
$S$ & number of MSA sequences \\
$R$ & number of residues \\
$c_m$ & MSA channel dimension \\
$c_z$ & Pair channel dimension \\
$H$ & number of attention heads \\
$d_h$ & per-head dimensionality (so $H d_h = \texttt{hidden\_dim}$) \\
\bottomrule
\end{tabular}
\end{center}

Each linear layer in the implementation corresponds to such a weight matrix $W$, 
with its shape determined by its input and output dimensions. For example,
\texttt{nn.Linear}$(c_m, 3 H d_h)$ produces
$W \in \mathbb{R}^{c_m \times 3H d_h}$, which is then split into \ac{QKV} projections in the row-attention module.

\subsection{Learned Parameters}\label{app:params}

The following weights are shared across the continuous-depth trajectory of the ODE function $f_\theta$:

\begin{center}
\begin{tabular}{@{}ll@{}}
\toprule
\textbf{Component} & \textbf{Learned weights} \\
\midrule
Row attention (MSA) & $W^{\text{row}}_{\mathrm{qkv}} \in \mathbb{R}^{c_m \times 3Hd_h}$, \;
$W^{\text{row}}_{\mathrm{gate}} \in \mathbb{R}^{c_m \times Hd_h}$, \;
$W^{\text{row}}_{\mathrm{out}} \in \mathbb{R}^{Hd_h \times c_m}$ \\
Pair bias & $W^{\text{pair-bias}} \in \mathbb{R}^{c_z \times H}$ \\
Column attention (MSA) & $W^{\text{col}}_{\mathrm{proj}} \in \mathbb{R}^{c_m \times Hd_h}$, \;
$W^{\text{col}}_{\mathrm{gate}} \in \mathbb{R}^{c_m \times Hd_h}$, \;
$W^{\text{col}}_{\mathrm{out}} \in \mathbb{R}^{Hd_h \times c_m}$ \\
MSA transition (FFN) & $W^{\text{msa}}_1 \in \mathbb{R}^{c_m \times 4c_m}$, \;
$W^{\text{msa}}_2 \in \mathbb{R}^{4c_m \times c_m}$ \\
Outer product mean & $W^a \in \mathbb{R}^{c_m \times 32}$, \;
$W^b \in \mathbb{R}^{c_m \times 32}$, \;
$W^{\text{opm}}_{\mathrm{out}} \in \mathbb{R}^{32 \times c_z}$ \\
Triangle update (Pair) & $W^{\triangle}_a \in \mathbb{R}^{c_z \times Hd_h}$, \;
$W^{\triangle}_b \in \mathbb{R}^{c_z \times Hd_h}$, \;
$W^{\triangle}_{\mathrm{gate}} \in \mathbb{R}^{c_z \times c_z}$, \;
$W^{\triangle}_{\mathrm{out}} \in \mathbb{R}^{Hd_h \times c_z}$ \\
Pair transition (FFN) & $W^{\text{pair}}_1 \in \mathbb{R}^{c_z \times 4c_z}$, \;
$W^{\text{pair}}_2 \in \mathbb{R}^{4c_z \times c_z}$ \\
Time embedding (\ac{MLP} on $t$) & $W^{t}_1 \in \mathbb{R}^{1 \times d_h}$, \;
$W^{t}_2 \in \mathbb{R}^{d_h \times 2}$ \; \\
\bottomrule
\end{tabular}
\end{center}

\begin{algorithm}[H]
\caption{Evoformer ODE vector field $f_\theta(m,z,t)$}\label{alg:ode-field}
\begin{algorithmic}[1]
\Require $m \in \mathbb{R}^{S \times R \times c_m}$, $z \in \mathbb{R}^{R \times R \times c_z}$, $t \in [0,1]$
\State $(\sigma_m, \sigma_z) \gets \textsc{TimeEmbedding}(t)$ \Comment{$\in (0,1)^2$}
\State $m \gets m + \textsc{MSA\_RowAttention\_WithPairBias}(m,z)$
\State $m \gets m + \textsc{MSA\_ColumnAttention}(m)$
\State $m \gets m + \textsc{MSA\_Transition}(m)$
\State $z \gets z + \textsc{OuterProductMean}(m)$
\State $z \gets z + \textsc{TriangleUpdate}(z)$
\State $z \gets z + \textsc{Pair\_Transition}(z)$
\State $\dot m \gets \sigma_m \cdot (\,m - \textbf{m}_{\mathrm{in}}\,)$ \Comment{broadcast $\sigma_m$ over $(S,R,c_m)$}
\State $\dot z \gets \sigma_z \cdot (\,z - \textbf{z}_{\mathrm{in}}\,)$ \Comment{broadcast $\sigma_z$ over $(R,R,c_z)$}
\State \Return $(\dot m, \dot z)$
\end{algorithmic}
\end{algorithm}

\noindent\emph{Notes.}
$\textbf{m}_{\mathrm{in}}$ and $\textbf{z}_{\mathrm{in}}$ denote the inputs to Algorithm~\ref{alg:ode-field} at the current solver time $t$ (they are not necessarily the values at $t{=}0$).
The ODE solver (\ac{RK4} in our experiments) integrates $(\dot m, \dot z)$ over $t \in [0,1]$ to produce the trajectory.

\begin{algorithm}[H]
\begin{spacing}{1.3}
\caption{\textsc{TimeEmbedding}$(t)$}\label{alg:time-embed}
\begin{algorithmic}[1]
\Require $t \in [0,1]$ \Comment{scalar depth variable}
\State $u \gets \mathrm{SiLU}(t W^{t}_1)\, W^{t}_2$
\State $(\sigma_m,\sigma_z) \gets \sigma(u)$ \Comment{$\sigma(\cdot)$ is sigmoid}
\State \Return $(\sigma_m,\sigma_z)$ \Comment{mixing coefficients}
\end{algorithmic}
\end{spacing}
\end{algorithm}

\begin{algorithm}[H]
\begin{spacing}{1.3}
\caption{\textsc{MSA\_RowAttention\_WithPairBias}$(m,z)$}\label{alg:row-attn}
\begin{algorithmic}[1]
\Require $m \in \mathbb{R}^{S\times R\times c_m}$,\; $z \in \mathbb{R}^{R\times R\times c_z}$
\State $\tilde m \gets \mathrm{LayerNorm}_{c_m}(m)$;\; $\tilde z \gets \mathrm{LayerNorm}_{c_z}(z)$
\State $(Q,K,V) \gets \text{split}\big(\tilde m W^{\text{row}}_{\mathrm{qkv}},\,3\big)$ \Comment{$\in \mathbb{R}^{S\times R\times H d_h}$}
\State reshape $Q,K,V$ to $(S,R,H,d_h)$
\State $B \gets (\tilde z W^{\text{pair-bias}}) \in \mathbb{R}^{R\times R\times H}$; broadcast to $(S,R,H,R)$
\State $\mathrm{Attn}(r,k,h) \gets \langle Q_{r,h},K_{k,h}\rangle/\sqrt{d_h} + B_{r,k,h}$
\State $\alpha \gets \mathrm{softmax}_k(\mathrm{Attn})$;\quad $O \gets \sum_k \alpha_{r,k,h} V_{k,h}$
\State $O \gets \mathrm{concat}_h(O)$ \Comment{$\in \mathbb{R}^{S\times R\times H d_h}$}
\State $g \gets \sigma(\tilde m W^{\text{row}}_{\mathrm{gate}})$
\State \Return $\,\big(g \odot O\big) W^{\text{row}}_{\mathrm{out}}$ \Comment{$\in \mathbb{R}^{S\times R\times c_m}$}
\end{algorithmic}
\end{spacing}
\end{algorithm}

\begin{algorithm}[H]
\begin{spacing}{1.3}
\caption{\textsc{MSA\_ColumnAttention}$(m)$}\label{alg:col-attn}
\begin{algorithmic}[1]
\Require $m \in \mathbb{R}^{S\times R\times c_m}$
\State $\tilde m \gets \mathrm{LayerNorm}_{c_m}(m)$
\State $U \gets \tilde m W^{\text{col}}_{\mathrm{proj}}$ \Comment{$\in \mathbb{R}^{S\times R\times H d_h}$}
\State $g \gets \sigma(\tilde m W^{\text{col}}_{\mathrm{gate}})$
\State \textbf{for} residue $r$ \textbf{do}\quad
$\alpha \gets \mathrm{softmax}_S(U_{:,r,:} U_{:,r,:}^\top / \sqrt{Hd_h})$;\;
$V \gets \alpha U_{:,r,:}$
\State \Return $\, \big(g \odot \mathrm{stack}_r(V)\big) W^{\text{col}}_{\mathrm{out}}$
\end{algorithmic}
\end{spacing}
\end{algorithm}

\begin{algorithm}[H]
\begin{spacing}{1.3}
\caption{\textsc{MSA\_Transition}$(m)$}\label{alg:msa-trans}
\begin{algorithmic}[1]
\Require $m \in \mathbb{R}^{S\times R\times c_m}$
\State $\tilde m \gets \mathrm{LayerNorm}_{c_m}(m)$
\State \Return $\, \mathrm{ReLU}(\tilde m W^{\text{msa}}_1)\, W^{\text{msa}}_2$
\end{algorithmic}
\end{spacing}
\end{algorithm}

\begin{algorithm}[H]
\begin{spacing}{1.3}
\caption{\textsc{OuterProductMean}$(m)$}\label{alg:opm}
\begin{algorithmic}[1]
\Require $m \in \mathbb{R}^{S\times R\times c_m}$
\State $\tilde m \gets \mathrm{LayerNorm}_{c_m}(m)$
\State $a \gets \tilde m W^a$;\;\; $b \gets \tilde m W^b$ \Comment{$\in \mathbb{R}^{S\times R\times 32}$}
\State $\mathrm{outer}_{s,i,j,:} \gets a_{s,i,:} \odot b_{s,j,:}$ \Comment{broadcast product}
\State $\overline{\mathrm{outer}}_{i,j,:} \gets \frac{1}{S}\sum_s \mathrm{outer}_{s,i,j,:}$
\State \Return $\overline{\mathrm{outer}}\, W^{\text{opm}}_{\mathrm{out}}$  \Comment{$\in \mathbb{R}^{R\times R\times c_z}$}
\end{algorithmic}
\end{spacing}
\end{algorithm}

\begin{algorithm}[H]
\begin{spacing}{1.3}
\caption{\textsc{TriangleUpdate}$(z)$}\label{alg:tri}
\begin{algorithmic}[1]
\Require $z \in \mathbb{R}^{R\times R\times c_z}$
\State $\tilde z \gets \mathrm{LayerNorm}_{c_z}(z)$
\State $A \gets \sigma(\tilde z W^{\triangle}_a)$ ;\;\;
$B \gets \sigma(\tilde z W^{\triangle}_b)$ \Comment{$\in \mathbb{R}^{R\times R\times H d_h}$}
\State $G \gets \sigma(\tilde z W^{\triangle}_{\mathrm{gate}})$ \Comment{$ \in \mathbb{R}^{R\times R\times c_z}$}
\State $T_{i,j,:} \gets \sum_k A_{i,k,:} \odot B_{k,j,:}$ \Comment{triangle mult., sum over $k$}
\State $T \gets \mathrm{LayerNorm}_{Hd_h}(T)$
\State \Return $\, G \odot (T W^{\triangle}_{\mathrm{out}})$ \Comment{$ \in \mathbb{R}^{R\times R\times c_z}$}
\end{algorithmic}
\end{spacing}
\end{algorithm}

\begin{algorithm}[H]
\begin{spacing}{1.3}
\caption{\textsc{Pair\_Transition}$(z)$}\label{alg:pair-trans}
\begin{algorithmic}[1]
\Require $z \in \mathbb{R}^{R\times R\times c_z}$
\State $\tilde z \gets \mathrm{LayerNorm}_{c_z}(z)$
\State \Return $\, \mathrm{ReLU}(\tilde z W^{\text{pair}}_1)\, W^{\text{pair}}_2$
\end{algorithmic}
\end{spacing}
\end{algorithm}

*All linear layers include learned bias vectors, omitted here for clarity

\end{document}